\documentclass{article}

\usepackage{microtype}
\usepackage{graphicx}
\usepackage{subfigure}
\usepackage{booktabs} % for professional tables

\usepackage{hyperref}

\usepackage[archival]{icml2023}

\usepackage{amsmath}
\usepackage{amssymb}
\usepackage{mathtools}
\usepackage{amsthm}

\usepackage[capitalize,noabbrev]{cleveref}
\usepackage{caption}

\theoremstyle{plain}
\newtheorem{theorem}{Theorem}[section]

\theoremstyle{definition}

\theoremstyle{remark}
\newtheorem{remark}[theorem]{Remark}

\usepackage[textsize=tiny]{todonotes}

\icmltitlerunning{An ML approach to Resolution of Singularities}

\newcommand{\CC}{{\mathbb{C}}}

\newcommand{\ZZ}{{\mathbb{Z}}}

\begin{document}

\twocolumn[
\icmltitle{An ML approach to resolution of singularities}



\icmlsetsymbol{equal}{*}

\begin{icmlauthorlist}
\icmlauthor{Gergely Bérczi}{equal,arh}
\icmlauthor{Honglu Fan}{equal,unige,eai}
\icmlauthor{Mingcong Zeng}{equal,mpi}
\end{icmlauthorlist}

\icmlaffiliation{arh}{Department of Mathematics, Aarhus University, Aarhus, Demark}
\icmlaffiliation{unige}{Department of Mathematics, University of Geneva, Geneva, Switzerland}
\icmlaffiliation{mpi}{MPIM Bonn, Bonn, Germany}
\icmlaffiliation{eai}{Eleuther AI}

\icmlcorrespondingauthor{Gergely Bérczi}{gergely.berczi@math.au.dk}
\icmlcorrespondingauthor{Honglu Fan}{honglu.fan@unige.ch}
\icmlcorrespondingauthor{Mingzong Zeng}{mingcongzeng@gmail.com}

\icmlkeywords{Machine Learning, ICML}

\vskip 0.3in
]



\printAffiliationsAndNotice{\icmlEqualContribution} % otherwise use the standard text.

\begin{abstract}
The solution set of a system of polynomial equations typically contains ill-behaved, singular points. Resolution is a fundamental process in geometry in which we replace singular points with smooth points, while keeping the rest of the solution set unchanged. Resolutions are not unique: the usual way to describe them involves repeatedly performing a fundamental operation known as "blowing-up", and the complexity of the resolution highly depends on certain choices.  The process can be translated into various versions of a 2-player game, the so-called Hironaka game, and a winning strategy for the first  player provides a solution to the resolution problem. In this paper we introduce a new approach to the Hironaka game that uses reinforcement learning agents to find optimal resolutions of singularities.
In certain domains, the trained model outperforms state-of-the-art selection heuristics in total number of polynomial additions performed, which provides a proof-of-concept that recent developments in machine learning have the potential to
improve performance of algorithms in symbolic
computation.
\end{abstract}

\section{Introduction}
Systems of multivariate polynomial equations, for instance
\begin{eqnarray*}
x^2z+yz^2+3y=0 \\ 
x^2yz-3=0
\end{eqnarray*}
play a central role in various scientific and engineering fields, as well
as geometry and topology. If the number of variables is $n$, then the solution set, which is also called variety in algebraic geometry, is a subset of $\mathbb{R}^n$ with rich geometry.  
The main technical challenge in solving these equations is the presence of ill-behaved, so-called singular points of the solution set. Geometrically, singularities can manifest as self-intersections, cusps, folds, or other irregularities in the shape of the solution set, see Appendix A for formal definition.  In applications, such as computer graphics, robotics, and computer vision, these irregularities can lead to visual artifacts, inaccurate simulations, or incorrect interpretations of data. When solving systems of polynomial equations, singularities can cause numerical instability. Near singular points, the equations can become ill-conditioned, leading to inaccuracies, divergence, or difficulty in finding reliable solutions. This is particularly relevant in applications where high precision and accuracy are required, such as scientific simulations or engineering design.

A fundamental problem in geometry about such systems is to remove these singular points by slightly modifying the solution set. This process is called resolution of singularities, and the main technical tool doing so is called blowing up.  For a toy example take one equation 
\[y^2-x^3+x^2=0\]
with a node (double point) singularity at $(0,0)$. To resolve this singularity, we substitute $y=xt$ to get $x^2(t^2-x-1)=0$, which has the green and black component. The non-singular blown-up curve is the black curve $t^2-x-1=0$. This resolution "blows up"  the origin: it replaces the origin 
with the projectivized tangent space of $\mathbb{R}^2$ at the origin. 
\begin{figure}[ht!]
\centering
%\captionsetup{justification=centering}
%\includegraphics[width=50mm]{Nodalcurve.png}
\includegraphics[width=40mm]{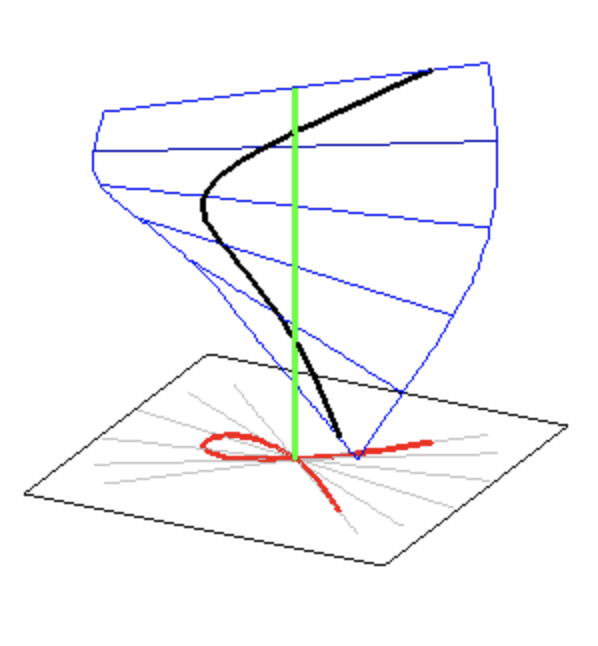}
\caption{A nodal singularity and its resolution. The red curve is $y^2-x^3+x^2=0$. The black curve is $t^2-x-1=0$, which is obtained as the main component by substituting $y=xt$ to get $x^2(t^2-x-1)=0$. The green line is the $x=0$ component in the equation $x^2(t^2-x-1)=0$.} \label{fig1}
\end{figure}

Resolution of singularities is an old, central problem in geometry with a long history.  Resolution of curves goes back to Newton, Riemann and Albanese, while resolution of surfaces has been extensively studied by the 19th century Italian algebraic geometry school. In 1964, Hironaka proved that resolution is possible for any singularity in characteristic $0$ \cite{hironaka}. This groundbreaking result re-defined the landscape of geometry and earned Hironaka the Fields Medal in 1970. While his initial proof was quite technical, subsequent algorithmic proofs \cite{vlo,atv} have been discovered with reduced complexity, resulting in resolution trees that outline the blow-up procedure. However, it is important to note that because the resolution of a singularity is not unique, discovering minimal resolutions remains a critical task in many mathematical fields. Interested readers can refer to \cite{kollar,abrICM} for more details. 

In this article we introduce a new approach that harnesses the power of deep reinforcement learning to seek out "good" solutions for resolving singularities. The idea starts with a classical observation by Hironaka \cite{hironaka3}: by playing a particular two-player game known as Hironaka's polyhedra game (see \ref{fundamental_hironaka_game}), one can obtain solutions that lead to local resolutions of some singularity types. However, it should be noted that this relationship does not hold in the reverse direction. We have identified two crucial observations in this regard: 
\begin{itemize}
    \item The Hironaka game is a Markov Decision Process(MDP) where actions only depend on the current game state (determined by the coordinates of discrete points in a space).
    \item The game can be generalized to resolutions of singularities under different constraints (resolving hypersurface singularities, only using weighted blow-ups, or even fully general resolutions such as \cite{hauser1})
\end{itemize}

The main focus of this article is to highlight that solutions to certain Markov Decision Processes (MDP) can be used to resolve specific singularities. Moreover, Reinforcement Learning is a commonly used technique to solve MDP.

The format of all such games is as follows: the game state consists of a finite set of points in either $\mathbb Z^n$ or $\mathbb Q^n$. Two players alternate turns and make decisions from a finite (but possibly extensive) range of actions that determine a linear transformation to be carried out on the space. After these transformations, a winning condition for the first player is checked (such as whether the Newton polytope  has only one vertex). The primary objective of the first player is to win the game in the fewest possible moves, while the second player's goal is to hinder the first player from winning by using adversarial actions. It should be noted that if the first player is not strategic enough, the game may continue indefinitely.

In general, the relationship between games and resolutions can be summarized as follows: the state of the game corresponds to a certain Newton polytope of the singularity on an affine chart. The first player chooses the blow-up center, and the second player chooses an affine chart on the blow-up. The linear transformation encodes how the Newton polytope changes under the blow-up using the transition of this affine chart.

On the math side, this work was motivated and subsequently guided by recent developments in the intersection theory of an important moduli space, the Hilbert scheme of points on manifolds \cite{bsz, berczitau}. Several  classical problems in enumerative geometry and mathematical physics can be reformulated using so-called tautological integrals over Hilbert schemes, and recent results show that the integral formula is not unique: any resolution tree of a certain singularity encodes a formula, which is a rational sum over the leaves of the tree. Hence the formula's complexity highly depends on the size of the resolution tree, and finding optimal trees using reinforcement learning is crucial in the analysis of the formula, which leads to new insights into these enumerative geometry questions. In particular, our random hitting host (described in \S \ref{randomhittinghost}) has provided a formula for a classical problem in topological enumerative geometry, see \S \ref{applications} for details.

%On the math side, this work is originally motivated and subsequently guided by the recent results  \cite{bsz,bercziThom,berczitau,berczitau2,berczitau3}. Using recently developed non-reductive geometric invariant theory (\cite{bkcoh}), a new integration formula over Hilbert scheme of points is developed, which in turn, gives new toric Thom polynomial formulas.  Many classical problems in enumerative geometry can be reformulated using these integrals. The integral formula is not unique: any resolution tree of a certain singularity encodes a formula, and hence its complexity highly depends on the size of the resolution tree. Finding optimal trees using reinforcement learning is crucial in the analysis of the formula, and it might help in proving old positivity conjectures of the subject (see Appendix for details)

On the machine learning side, the work is inspired by recent breakthrough results \cite{davies2021} in ML-assisted proofs in pure mathematical problems, and by the deep reinforcement learning techniques which has been a powerful tool to solve problems that can be phrased as (or close to) a Markov decision process \cite{buchberger}. With the success of AlphaZero (\cite{alphazero}), the deep reinforcement learning is further amplified by the power of planning using tree search when the rules of the environment is perfectly understood. By connecting with the problem of resolution of singularities, we hope to provide deep reinforcement learning an additional use case that has a broad impact in mathematics with its own special challenges.

Deep reinforcement learning has recently shown potential in various mathematical domains where computation is a crucial aspect. One such instance is the application of deep reinforcement learning to Buchberger's algorithm \cite{buchberger}, which was proposed by Dylan Peifer, Michael Stillman, and Daniel Halpern-Leistner. Their approach employs reinforcement learning agents for S-pair selection, which is a process involving a two-player game, similar to the Hironaka game we discussed.

This article, while heavily focusing on the mathematical motivations and the mathematics-reinforcement learning translations, aims to also show some preliminary experiments suggesting the feasibility of applying deep reinforcement learning in the family of singularity resolution problems. 

%The policies we have trained are not yet enough to obtain deep mathematical insights, but further improving and scaling the RL algorithm is still an ongoing project and it is left for future articles.

\noindent \textbf{Acknowledgement}
The first author is indebted to the organisers and  members of the "Machine Learning for the Working Mathematician" seminar organised by Joel Gibson, Georg Gottwald, and Geordie Williamson at University of Sydney \cite{mlwm}  We are also deeply indepted to High Flyer AI and Google TRC program for the computing support.  
We also greatly appreciate the Google Cloud Research Credits we received for using the Google Cloud Platform.

%Note: the first author learned about this work, along with the DeepMind AlphaZero project, and \cite{davies2021} at G. Williamson's Machine Learning for the Working Mathematician seminar in Spring 2022. 

%Our project has grown beyond the original motivations in enumerative geometry, and we found further, potentially far-reaching applications of the RL approach, collected in the Appendix. The ultimate goal of our project is to study the complexity of resolution trees, and find optimal resolutions using Reinforcement Learning. We calculate invariants of singularities from their resolution trees and attack the classification problem by comparing singularities and understanding their hierarchies based on their resolution trees. 

%Our results are collected in this  \href{https://github.com/honglu2875/hironaka}{Github repo}.

\section{A fundamental version of the game}\label{fundamental_hironaka_game}
All of the games discussed in this paper can be regarded as variations of one fundamental, simple version. The two players play asymmetric roles, and for convenience we call Player $1$ the ``host", and Player $2$ the ``agent". The rules of this basic version are the following:

A state is represented by a finite set of lattice points 
\[S\subset \mathbb Z_+^n=\{(x_1,\ldots, x_n):x_i>0 \text{ integer for } 1\le i \le n\}.\]

\begin{itemize}
    \item the host chooses a subset $I\subset \{1,2,\cdots, n\}$ such that $|I|\geq 2$, and 
    \item the agent chooses a number $i\in I$.
\end{itemize}

The selected pair $(I,i)$ deteremines the following linear change of variables:

$$T_{I,i}: x_j \mapsto \begin{cases}x_j, &\qquad\text{if } i\neq j \\ \sum\limits_{k\in I} x_k, &\qquad\text{if }i=j
\end{cases},$$

After $S$ is transformed into $T_{I,i}(S) \subset \ZZ_+^n$, we remove all points sitting in the interior of the Newton polygon spanned by $T_{I,i}(S)$. That is, the new state is $S'=T_{I,i}(S) \setminus N(T_{I,i}(S))$ where 
\begin{equation*}
\begin{split}
    &N(T_{I,i}(S)) \\
    = & \{(x_1,\cdots,x_n)\in T_{I,i}(S): \exists (x_1',\cdots,x_n')\in S' \\
    &\text{ such that } x_i\leq x'_i,~ \forall 1\le i \le n\}.
\end{split}
\end{equation*}
A state is terminal if it consists of one single point. In this case, the game will not continue and the host is the winner. The host can also have an incentive to terminate the game in the fewest possible steps, and the resulting solutions can have significant mathematical implications.

A host with a winning strategy against all possible agents corresponds to a resolution of the singularity corresponding to the initial state of the game. An example can be seen in \href{https://www.youtube.com/watch?v=s2LYtd_UPY8}{this video}. To connect with our context, for example, the E8 singularity corresponds to games with the initial state of $\{(2,0,0), (0,3,0), (0,0,5)\}$.

\begin{algorithm}[tb]
   \caption{The Hironaka game}
   \label{alg:example}
\begin{algorithmic}
   \STATE {\bfseries Input:} Finite set $S\subset \ZZ_+^n$
   \STATE {\bfseries while} Newton polygon of $S$ has $\ge 2$ vertices \textbf{do:}
   
    \textbf{\ \ \ Player 1:} Choose subset $I\subset \{1,\ldots, n\}$ with $|I|\ge 2$
    
    \textbf{\ \ \ Player 2:} Choose an element $i\in I$
    
    \textbf{\ \ \ Transformation:} $T_{I,i}: \ZZ_+^n \to \ZZ_+^n$; new set is $T_{I,i}(S)$.
\end{algorithmic}
\end{algorithm}

The game does not necessarily terminate in finite number of steps. But the existence of a winning strategy is implied by Hironaka's Fields medal result \cite{hironaka}. However, the winning strategy is executed and the quality of the strategy (e.g., minimal amount of game steps against the smartest agents) still concern many directions in algebraic geometry.

\begin{remark}
Although the original Hironaka game does not contain this step, we can reduce the length of the game (and hence the blow-up process) by a simple additional step when we form the new state $S'$ from $S$. We call this the translation step: for a set of points $Z \subset \ZZ_+^n$ let 
\[z_i^{\min}=\min_{(z_1,\ldots, z_n)\in Z} z_i\]
denote the minimal $i$th coordinate in $Z$, and $z^{\min}=(z_1^{\min},\ldots, z_n^{\min})$. Then the shifted set is  
\[Z^{sh}=\{z-z^{\min}: z\in Z\}\]
and the modified rule is that the new state in the Hironaka game is 
\[S'=(T_{I,i}(S) \setminus N(T_{I,i}(S)))^{sh}.\]
Geometrically, this corresponds to removing exceptional divisors, and keeping only the strict transform of the singularity. In Figure \ref{fig1} the strict transform is the black (nonsingular) blown-up curve.
\end{remark}

%{\color{blue} Add geometry of the state change using Newton polytope. Don't we want to divide all monomials with their hcd in the state change, i.e translate the points towards the origin?}
%{\color{red} The original Hironaka game doesn't shift towards the origin. I don't have a good explanation of his original motivation and why he does not shift. Do you have a good explanation or remark for this? For the geometric examples, I already added the shift when training hosts for all the examples I demonstrated in Appendix C. I already explained it in the second paragraph of Appendix C. Please take a look.}

\begin{remark} We make two observations on the rules. 
\begin{enumerate}
    \item As one might have already noticed from the rules, the removal of interior points of the Newton polytope is in fact not necessary. One may simply carry on without removal of any point, and only draw the Newton polytope to declare the terminal state if all but one points are in the interior. This observation is helpful in avoiding our discussions becoming lengthy when we define its Markov Decision Process later.
    \item In this particular version of the game, we can freely scale the points in $\mathbb Z^n$ by a common scalar. This allows for a few equivalent implementations including bounding the game states inside a unit sphere.
\end{enumerate}
\end{remark}

\section{Reinforcement learning}

Reinforcement learning (RL) has recently shown remarkable success in a variety of applications, from playing games to robotic control and natural language processing. This, in principle, also applies to our mathematical problem which can be phrased as a Markov Decision Problem (MDP).

We focus on two popular RL algorithms, Monte Carlo tree search (MCTS) and Deep Q-Networks (DQN). MCTS is a planning algorithm that searches the tree of possible actions to find the best sequence of actions that leads to the optimal solution. DQN, on the other hand, is a value-based algorithm that uses a deep neural network to approximate the action-value function and update the policy. In addition to MCTS and DQN, Proximal Policy Optimization (PPO) is a popular on-policy RL algorithm that has shown promising results in several applications. However, in our study, we only used MCTS and DQN for solving the optimization problem.

Some examples of the influential applications in the field of RL include [1] for DQN, [2] for MCTS, and [3] for PPO.

While our experimental results are not able to deterministically solve all singularities of a specifical dimension, they demonstrate the feasibility of using MCTS and DQN for solving Hironaka-style games and similar optimization problems. We observe improvements in the policy during training, which suggests that further refinement and exploration of RL techniques could lead to better performance. We will highlights the challenges and opportunities of applying RL techniques to mathematical optimization problems and provides a valuable contribution to the field of RL. We hope that our work will inspire further research in this direction and pave the way for more successful applications of RL in solving optimization problems.

\subsection{The asymmetric objectives}
Just like most RL setups, we have a set of game states $\mathfrak S$ (in the basic Hironaka game: $\mathfrak S = (\ZZ^n)^k$ where $k$ is the number of points in a given initial state). The rules of the games can be phrased as MDPs (section \ref{fundamental_hironaka_game}), and the two players can have their rewards functions to optimize (will be detailed in section \ref{mdp}). 

But a subtle point in the translation between mathematical setup and the RL environment is the following.
\begin{quote}
A resolution of a singularity corresponds to {\bf a fixed and deterministic policy} of the first player on {\bf all} states reachable from a given initial state $S$. In particular, a full policy of the first player is determined by its reactions upon all possible actions from the second player.
\end{quote}
This suggests that the second player is only an adversarial challenger that is allowed to freely explore/retry and find the best way to delay the game from ending.

So, the overall goal is slightly different than optimizing a single pair of policies: we view a fixed resolution of singularity as a ``host" of the game against {\bf all possible} adversarial agents. They have asymmetric objectives:
\begin{itemize}
    \item An adversarial agent needs to delay the games of one host for as long as possible.
    \item But the best host needs to be able to terminate in finite steps against all agents, or even minimize the amount of steps against the best adversarial agent.
\end{itemize}
This is also why in section \ref{fundamental_hironaka_game}, we call the first player the ``host", and the second player the ``agent". 

\subsection{The Markov Decision Processes}\label{mdp}

\subsubsection{The training setups of the 2-player problem}
More formally, the basic Hironaka game can be converted into Markov Decision
Process (MDP) if one of the players is fixed. As mentioned before, the process this article adopts goes into two steps:
\begin{enumerate}
    \item Fix a host to search for the best adversarial agent;
    \item Fix one or more adversarial agent with top performances and search for a better host policy.
\end{enumerate}
And the training will be continuously iterating through these two processes.

We also note that there are two different ways to set up this two-player problem.

\begin{itemize}
    \item One may unify the host's and the agent's observation space, action space and reward space by leaving placeholders in their definitions of states and actions. It will turn into a symmetric game that can be improved by self-play.
    \item One may fix a pool of host policies and adversarial agents, simulate the playoffs and select the elites using evolution methods.
\end{itemize}

The first view is especially useful when incorporating planning (e.g. using MCTS) into the RL setup. Whether to improve by self-play will influence implementation details, and it is not yet clear which one is better.

The second view is rather interesting from research perspectives. It requires a formulation of rewards that aggregates across the whole pool of fixed agents. One can only apply the group-evaluation idea only as an evaluation metric (Elo rating, etc.), or go with full-scale evolution experiments in search for strategies that are universally good. These are very involved directions. We have made some preliminary experiments of maintaining dynamic pools of DQN agents and select players using MAP-Elites, but the results are not yet enough for a systematic presentation. We leave them for future research.

For this article, we do not go into details about these two alternatives. We focus on the iterations of the 2-step process fixing agents and hosts in turn.

\subsubsection{Searching for the best adversarial agent}
The first process is to fix a host in search for the best adversarial agent policy. We define the MDP as follows.

At a time step $t$, let $S_t \subset \mathbb (Z^n)^k$ be the current state of the game. For a fixed host policy, a choice of coordinates $I_t\subset \{1,2,\cdots,n\}$ will be made along with a given game state. The agent’s observation $s_t = (S_t, I_t)$ is such a pair consisting of the game state and the host's choice. 

Strictly speaking, the action space of the agent is exactly the finite set $I_t$. But to fit the definition of MDP, we can simply define the action space $A$ to be the full $\{1,\cdots,n\}$ and impose a probability $0$ for actions outside $I_t$.

The reward $R_{a_t}(s_t, s_{t+1})$ for a given action $a_t\in A$ can be simply defined as $0$ if the game state $S_{t+1}$ is not terminal, and $-1$ if $S_{t+1}$ is terminal.

%{\color{blue} Naive question: Why exactly this reward function? Would e.g +1,-100 significantly change the training process?}
%{\color{red} Yes it will, and it might not change the training in the way we want it. Rewards are coefficients in a giant recursion (see the optimization objective below). There is no good way to say what coefficients lead to what. There might even be some chamber structure about the convergence of that recursion.

%I choose this reward because it is standard. In AlphaGo, AlphaZero, the reward is also 0 if not ended, 1 if win, -1 if lose.}

As a common practice, we look for a policy $\pi$ which determines the actions by $a_t=\pi(s_t)$, optimizing the following objective:
\[\mathbb E\left[\sum\limits_{t=0}^\infty \gamma^t R_{a_t}(s_t, s_{t+1})\right],\]
where $\gamma$ is a discount factor between $0$ and $1$, $\mathbb E$ is the mathematical expectation over all transitions. By having this discounted sum, it also encourages the agent to delay the end the game (if at all) as much as possible. Since we are working on a deterministic MDP with deterministic policies, the $\mathbb E$ becomes redundant for a fixed initial state and the host policy. Thus, following the definitions, we are performing a search algorithm over agent actions that optimizes the discounted sum of rewards.

\subsubsection{Searching for the best host}\label{searching_for_host}
Once a strong agent policy is found, we in turn fix the agent policy, and look for a good host policy to counter that. The MDP is very similar except that the state $s_t$ consists of $S_t$ only, and the action space $a_t$ ranges over all subsets $I\subset\{1,\cdots,n\}$ with more than $1$ element. The reward $R'_{a_t}(s_t, s_{t+1})$ is $0$ if $s_{t+1}$ does not terminate, and $1$ if $s_{t+1}$ terminates.

A potential risk of this iterative approach is that the evolution of host-agent pair might get stuck in loops countering each other without achieving high performances over all counter-parties. Therefore, at least theoretically, the optimizing objective for host player 
\begin{equation}\label{host-obj}
\mathbb E_{\pi}\left[\sum\limits_{t=0}^\infty\gamma^t R'_{a_t}(s_t, s_{t+1})\right]
\end{equation}
must average over all agent policies $\pi$.

In practice, evaluating Equation \eqref{host-obj} is often impractical. What we do is to simply fix a collection of agent policies $\mathcal P$, and average over them:
\begin{equation}\label{practical-obj}
\dfrac{1}{|\mathcal P|}\sum\limits_{\pi\in\mathcal P}\left[\sum\limits_{t=0}^\infty \gamma^t R_{a_t}(s_t, s_{t+1})\right].
\end{equation}
Although a host is trained against a limited number of agents, generalizability is observed and will be demonstrated later in Figure \ref{host-choosefirst-random}.

\subsection{A simple metric}\label{rho-metric}

We introduce a metric which is helpful in measuring the performance of the host and agent during evaluations. The metric is based on the idea that for a fixed amount of time steps, a smart host should be able to play more games while a smart agent should be able to play fewer games. On the host side this means more low-turn solutions while on the agent side this means that more games are elongated in terms of steps, induced by the competent adversary.

Given a bounded subset $V\subset \mathbb (Z^n)^k$ of initial states and a fixed host-agent pair, a process of continuous game-play is run for a fixed number of steps $m$, with an immediate restart after terminal state by uniformly sampling another initial state from $V$. Denote such an $m$-step game sequence by $G_m$. The metric is a simple ratio between the number of games played and the number of steps taken:
\[\rho_{V,m} = \mathbb E_{G_m}(\rho_{G_m}(m))=\frac{\mathbb E_{G_m}(g_{G_m}(m))}{m}\]
where $\mathbb E_{G_m}$ is the mathematical expectation over all game sequences $G_m$ and $g(m)$ is the number of different games during these $m$ steps. If a limit
\[\rho_V=\lim_{m\to \infty}\mathbb E_{G_m}(\rho_{G_m}(m))\]
exists, it could be a good measure for the pair of host and agent: If $\rho_V$ is high, the host policy is uniformly stronger than the agent, vice versa.
Our tests suggest the existence of the limit, but for a formal proof significantly more formality linking the proofs of the Hironaka theorem to our MDP formulation is needed. Our experiments show that 
\begin{enumerate}
    \item empirically $\rho_{V, m}$ is observed to have a tendency of convergence after sampling enough games and let $m$ grow;
    \item in practice, we fix a uniform and sufficiently large $m$ and only use $\rho_{V, m}$ as the metric due to efficiency.
\end{enumerate} 

For the rest of this paper, for each experiment we fix a sufficiently large integer $N$, and we sample initial states from the set \[V=\{(x_1, \cdots, x_n) ~|~ 1\leq x_i\leq N, \forall 1\leq i\leq n\}^k.\]
For notation convenience, we later drop the notion $V$ in $\rho_{V, m}, \rho_V$ and only denote them by $\rho_m$ and $\rho$.

Although it is very useful in evaluation and selection of policies, there are a couple limitations of this metric:
\begin{enumerate}
    \item There are subtle differences between $\rho_V$ and our mathematical goal: look for the host policy that consistently generates blow-ups that require as few steps as possible in \emph{all} charts (i.e., against all agent policies).
    \item The limit is not proven to exist, and estimating it requires many sampling from random games.
\end{enumerate}

Nevertheless, by using this empirical metric, we are able to select strong policies. For example, the one showcased in Appendix \ref{appendix:alphazero} is selected among a sequence of checkpoints using its empirical $\rho$ value against two very basic hard-coded agent benchmark (choose-first agent and choose-last agent).

\subsection{Agent benchmarks and generalizability}
Back to \eqref{practical-obj}, since we sample a set of agents, the simplest scenario is when they are fixed policies throughout the training. We have the following simple benchmarks.

\subsubsection{Constant agents and random agents}
There are two obvious choices to hard-code agent strategies: the constant policy and the random policy. 

For the constant policy, we focus on the special case where the agent always choose the first available coordinate from the host's choice $I\subset\{1,\cdots,n\}$, ordered according to the coordinate labels. We call it the choose-first agent for short.

Choose-first agent is in fact one of the strongest hard-coded policy we have in terms of the empirical metric $\rho$ introduced in section \ref{rho-metric}. This obviously generalizes to all agents who always choose a fixed action such as ``choose-last". 

\subsubsection{Generalizability on different agents}
Since in practice, we sample a limited number of agents in the objective \eqref{practical-obj}, it is important to ask whether the training against one agent can be generalized to another.
We observe that hosts trained against constant policies can already generalize its performance to other policies.

\begin{figure}[ht!]
\centering
\includegraphics[width=85mm]{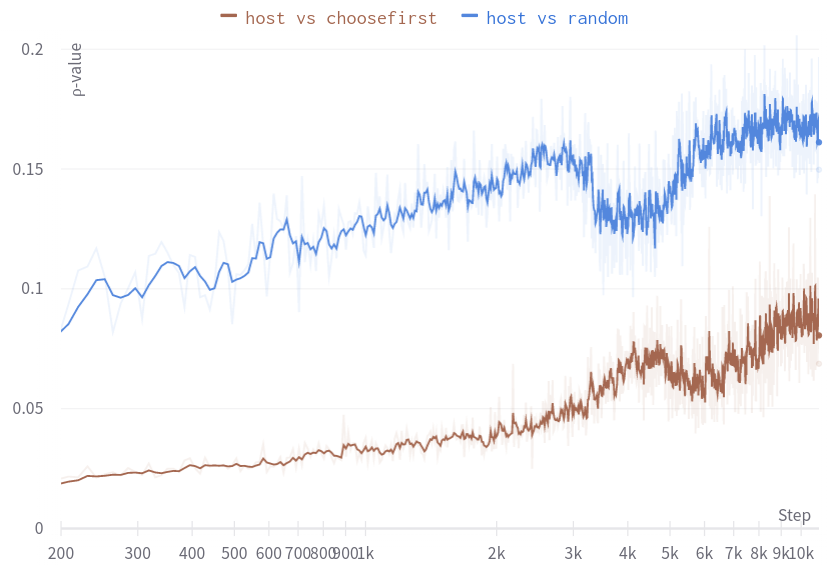} 
\caption{Host trained against the choose-first agent and the random agent} \label{host-choosefirst-random}
\end{figure}

An example is shown in Figure \ref{host-choosefirst-random}. The host is a 2-layer residual network with a $256$ hidden dimension trained against the fixed choose-first agent using double DQN (\cite{ddqn}). The plot includes the evaluations of the same host network against the choose-first agent and the random agent throughout its training process. The y-axis are approximated by $\rho_n$ with $n=1000$ and use log scales. The x-axis starts at $200$-step for warm-ups. We observe the following:
\begin{itemize}
    \item The host vs choose-first curve in Figure \ref{host-choosefirst-random} shows an expected improvement throughout training, as the choose-first agent is directly used in the roll-outs as the fixed adversarial agent.
    \item The host vs random curve in Figure \ref{host-choosefirst-random} shows a correlated (though not fully in sync) trajectory of improvements against random agent policy against whom the host policy has never played.
\end{itemize}
 . 

\subsection{Host benchmarks}
Other than a learned host network, there are the following host strategies coming from or inspired by the mathematical literatures.
\subsubsection{The Zeillinger host}\label{zeillinger-host}
We start with the simplest policy that is guaranteed to terminate in finite steps due to Zeillinger \cite{zeillinger}.

Let $N$ be the Newton polyhedron of the finite set of points $S\in\mathbb Z^n$. Zeillinger proved that the host wins the polyhedra game if he uses the following strategy:
\begin{enumerate}
    \item If $N$ is an orthant, the game is already won.
    \item If $N$ is not an orthant, choose
$I=\{k,l\} \subset \{1,\ldots n\}$
with $1 \le k <l \le n$ such that $w_k$, is a minimal and $w_l$ is a maximal component of a characteristic vector $(w_1,\ldots, w_n)$ of $N$.
\end{enumerate} 
The definition of the characteristic vector is a bit technical, but can be found in \cite{zeillinger}. The main point is that the Zeillinger host always picks a pair of coordinates. Although it is a winning strategy, the length of the game against any agent--and hence the size of the resolution tree--is large. Zeillinger's strategy limits each choice to a finite number of pairs $(k,l)$ any of which ends the game in finite steps. But since the choices are not unique, there might be a number of choices at each step resulting in different depths of the trees. 

\subsubsection{The Spivakovsky host} 
The first known strategy with guaranteed finite termination predates Zeillinger and was given by Spivakovsky \cite{spivakovsky2}. The Spivakovsky host works with coordinate subsets $I \subset \{1,\ldots, n\}$ which are hitting sets: we call $I\subset \{1,\ldots, n\}$ a hitting set if at least one coordinate in $I$ is nonzero for all vertices of the Newton polygon $N$. We leave the exact technical details of how $I$ is selected for \cite{zeillinger,spivakovsky2}, but we point out two key features: 
\begin{enumerate}
    \item At each step the Spivakovsky host picks a hitting set of the Newton polygon $N$. 
    \item The Spivakovsky host works with big hitting sets, and hence it is relatively slow, the game is long. 
    \end{enumerate}
\subsubsection{The Random Hitting host}\label{randomhittinghost}
Our hand-made calculations with Thom polynomials for a slightly modified game (the Thom game) suggest that for an optimal (i.e minimal) resolution tree the host should pick a hitting set at each step, and the size of this should be small. %The resolution trees of \cite{bercziThom} are conjecturally close to the minimal resolutions. 
However, deterministic selection of a minimal hitting set, such as constant selection policy, does not work in general: one can stuck in an infinite loop of transformations. Hence we introduced the following Random Hitting host: 
\begin{enumerate}
    \item The host picks a minimial hitting set at each step 
    \item The choice is random among the hitting sets of minimal size. 
\end{enumerate}
Based on a large number of examples, the Random Hitting host performs the best. A future direction of research will be to find the best performing host who works with minimal hitting sets, using a reinforcement learning model.

\subsection{Planning with tree search}\label{MCTS_planning}
Our strongest performer is obtained through AlphaZero-style RL training with MCTS planning (see for example, \cite{alphazero}). The idea of mixing deep learning and MCTS planning is to use the policy network as a heuristic to guide the Monte Carlo tree search in order to determine the action distribution of a state. The Monte Carlo tree search in turn provides an improved policy comparing to the vanilla result from the policy network. Using the MCTS-improved policies, the subsequent self-plays are collected and used to improve the policy network in a supervised manner. The key point is that this feedback loop achieves policy improvements in an overall unsupervised way. But this is in practice difficult to stabilize, and it requires efforts to experiment the model architectures and tune the hyper-parameters.

We approximate the $\rho$ metric mentioned in section \ref{rho-metric} and summarize in Table \ref{eval}.
\begin{table}[t]
\caption{$\rho$-evaluation.}
\label{eval}
\vskip 0.1in
\begin{center}
\begin{small}
%\begin{sc}
    \begin{tabular}{l|ccc}
    %\backslashbox{host}{agent}
         & random & choose-first & choose-last \\
         \midrule
         MCTS & 0.215 & 0.126 & 0.115 \\
         Choose-all & 0.253 & 0.032 & 0.032 \\
         Zeillinger & 0.232 & 0.157 & 0.161 \\
         \bottomrule
    \end{tabular}
%\end{sc}
\end{small}
\end{center}
\vskip -0.1in
\end{table}

The MCTS host is trained for surface singularities (states coming from $\mathbb Z^3$) with a maximum of $20$ points. The host is trained against an agent network that also improves using MCTS. We perform the training back-and-forth by first fixing the agent policy, then fixing the host policy, and then back fixing the agent policy, etc. 
For more training details, we refer to Appendix \ref{appendix:alphazero}.

In Table \ref{eval},
\begin{itemize}
    \item ``MCTS" is our policy guided by the trained network and perform MCTS for $100$ steps.
    \item ``Choose-all" is the constant host policy that picks the full set of coordinate in every single move.
    \item ``Zeillinger" uses the strategy in section \ref{zeillinger-host}. It is theoretically proven to be a winning strategy against all possible adversarial agents.
\end{itemize}

In the Appendix \ref{appendix:alphazero}, we demonstrate this MCTS host policy on a few mathematical meaningful examples (du Val singularities).

There are two note-worthy features about this host policy:
\begin{itemize}
    \item most of the host policies (including those with high $\rho$ trained from direct applications of DQN, PPO, etc.) are almost never capable of ending the game in finite steps against {\bf all} possible agent policy. This problem occurs even for initial states with as few as $2$-$3$ points. But our MCTS host is demonstrated to end all the initial states corresponding to du Val singularities in finite steps against all possible agents.
    \item Although Zeillinger host always guarantees a resolution, it always blows up codimension-$2$ strata and in a lot of cases almost never a minimal resolution. But in Appendix \ref{appendix:alphazero}, we show that our host is able to provide minimal or close-to-minimal resolutions for some du Val singularities. 
\end{itemize}

This shows the feasibility of applying deep RL to finding useful host policies corresponding to valid or even minimal resolutions of singularities. We only tried the most basic implementations and there are still a wide range of possible improvements. We are still exploring it in an ongoing research.

\section{Applications and conclusions}\label{applications}

While the existence of resolutions of singularities is proved (in characteristic zero), the development of effective algorithms for resolving singularities remains a challenge. Constructing explicit resolutions can be computationally demanding and often involves deep geometric and combinatorial techniques. Developing efficient algorithms that work in general settings is an ongoing research area. Our work offers evidence for the relevance of an ML-approach towards two main challenges: 

\textbf{Minimal resolutions:} Given that multiple resolutions of a singular variety may exist, a natural question is whether there exists a notion of "minimal resolution." Minimal resolutions capture the essential geometric and arithmetic properties of the singularity and provide a canonical representation. Understanding the existence and uniqueness of minimal resolutions is an active research topic.

\textbf{Classification of singularities:} While there are some well-understood classes of singularities, such as ordinary double points or rational singularities, a complete classification of singularities is still an open problem. Developing a comprehensive classification scheme by extracting patterns of resolution trees using ML would contribute to a deeper understanding of singularities and their resolutions.

\subsection{Topology of maps with an outlook}

This work was originally motivated by the first authors' pure math paper \cite{bercziThom} on the topology of maps, hence in this final section we collect applications which are closer to the authors' expertise. 
Global singularity theory is a classical subject in geometry which classifies singularities of maps $f:\CC^n \to \CC^m$, and 
describes topological reasons for their appearance. One of its central questions is to determine the (cohomological) locus where $f$ has a given type of singularity. This (cohomology) locus is called the Thom polynomial of the singularity, named after Ren\'e Thom, who introduced and studied them in the 1950's.

In \cite{bsz} a formula for Thom polynomials of Morin singularities was developed, which can be reduced to toric geometry, in particular, to a sum of rational expressions over leaves of the blow-up tree obtained by a variant of the Hironaka game (which we call the Thom game) on a special singularity. The formula has the form 
%can be informally summarised as follows. 
%\begin{theorem}\label{tpformula} For arbitrary integers $k\ll n \le m$ the Thom polynomial for the Morin singularity $A_k$ with
%$n$-dimensional source space and $m$-dimensional target space is given by an iterated residue formula
\[\mathrm{Tp}_k^{n,m}=\operatornamewithlimits{\mathrm{Res}}\limits_{\mathbf{z}=\infty}\left(\sum_{L \in \mathcal{T}_k} R_{L}(\mathbf{z})\right) \prod_{i=1}^k c_f(1/z_i)d\mathbf{z}\]
where: $\mathbf{z}=(z_1,\ldots, z_k)$ are the residue variables, and the iterated residue is the coefficient of $(z_1 \ldots z_k)^{-1}$ after we expand the rational expression on the contour $z_1 \ll \ldots \ll z_k$;  $\mathcal{T}_k$ is the set of leaves of an (arbitrary) blow-up tree of a certain Thom ideal $I_k$; $R_L$ is a rational expression assigned to the label $L$; $c_f(1/z_i)$ stands for a generating function of cohomology classes for the map $f:\CC^n \to \CC^m$ (these are called Chern classes).
    % which is defined as \begin{equation*}
    %    \frac{1+\sum\limits_{i=1}^m %c_i(TM)}{1+\sum\limits_{i=1}^n c_i(TN)}
    %    =1+c_1(TM-TN)+\ldots .
    %\end{equation*}
%\end{itemize}
%\end{theorem}

Using a modified resolution game, the Thom game (see Appendix), we managed to construct a blow-up tree $\mathcal{T}_k$ in the theorem for $k\le 7$. 
The complexity of the formula is determined by the complexity of the Thom tree, but unfortunately, these resolution trees are quickly becoming oversized as $k$ increases, and finding small resolution trees is crucial in understanding the structure and symmetries of the formula. Our formula for $k=7$ is a new result: we only knew Thom polynomials before  up to $k=6$.

We believe our approach has big potential in tackling other classical questions in enumerative geometry, such as
\begin{itemize}\itemsep-0.2em
\item The Chern positivity conjecture of Thom polynomials. Rim\'anyi \cite{rimanyi} conjectured that the Thom polynomials expressed in the Chern classes of $f$ have nonnegative integer 
coefficients. This conjecture remained hopeless and intact since its formulation. 
%It is a topological analogue of the 
%Stanley e-positivity conjecture for chromatic polynomials on graphs. 

%We hope that using RL and deep networks to find optimal resolution trees for Thom ideals might help us in understanding these nonnegative coefficients, and helps us in identifying the "worst" coefficients. 

%These include some of the fundamental questions in enumerative topology and geometry:
\item Counting plane curves with given set of singularities.
\item Counting maps between manifolds with given set of singularities
\item Determining (cohomological) locus of maps where the map has given singularities. 
\item Conjectures on integrals in mathematical physics (Segre-Verlinde duality)
\end{itemize}

%An iterated residue formula, which is similar to the one we presented in Theorem \ref{tpformula}, can be developed for tautological integrals over Hilbert scheme of points, 
%This involves a similar blow-up tree as for Thom polynomials, generated by the Thom game, and it is encoded by a sum of certain rational expressions over the leaves of the tree. Hence the complexity of the formula is again determined by the complexity of the Thom tree.  

%\subsection{The Monomialisation Conjecture of morphisms}

%The Monomialization Conjecture is a relative formulation of the resolution problem, and this is a massive open question.  It asserts that any morphism of varieties 
%over a field of characteristic zero can be transformed via blowups of source and target into a monomial morphism, 
%i.e., a morphism which can be expressed in suitable local coordinates by monomials.
%There have been important recent advances by Cutkosky and Abramovich, Karu, Matsuki and Wlodarcyzk. 
%In positive characteristic there are simple counterexamples.

\bibliography{bib}
\bibliographystyle{icml2023}

\newpage
\appendix
\onecolumn
\section{Appendix: A brief mathematical background}
In 1964 Hironaka proved that it was possible to resolve singularities of varieties over fields of characteristic 0 by repeatedly blowing up along non-singular subvarieties, using a very complicated argument by induction on the dimension. Over the last 60 years several other proofs were discovered, with reduced complexity, including Bierstone \& Milman \cite{bierstone}, Encinas \& Villamayor \cite{ev98}, Wlodarczyk \cite{vlo}, McQuillen \cite{quillen} and Abramovich \& Tempkin \& Wlodarchyk \cite{atv}.

A resolution can be described by a series of blowing-ups, and these elementary operations can be arranged into a blow-up graph, which is a rooted tree labelled by clusters of variables. This tree is not unique; its size and complexity highly depend on some choices. Our knowledge about this complexity is very limited. To study the complexity of resolution trees, and to find optimal resolutions using reinforcement learning is the ultimate goal of what we propose in this paper.

\subsection{What is a resolution of singularity}

An affine algebraic variety  
$$X =\{(x_1,\ldots, x_n): f_1(x_1,\ldots, x_n)=\ldots =f_k(x_1,\ldots, x_n)=0\} \subset \mathbb{A}^n$$
is the common zero locus of polynomial equations. Affine varieties play central role in mathematics, physics and biology.   
Affine varieties cut out by one polynomial equation are called affine hypersurfaces. E.g
$$X=\{(x_1,\ldots, x_n):f(x_1,\ldots, x_n)=0\}$$

\subsection{Singularities}

We can think of varieties as "shapes in affine spaces", and at a generic point 
$x \in X$ the variety locally is $\mathbb{A}^r$ for some $r$, which we call the dimension of $X$.
However, there are special, ill-behaved points, where the local geometry of $X$ is less patent.

The affine variety $X$ is singular at a point $a \in X$ if the Jacobian matrix
$$\mathrm{Jac}(X,a)=\left(\frac{\partial f_i}{\partial x_j}\right)(a)$$
at a is of rank smaller than $n-\text{dim}(X)$. The set of singular points of $X$ is called the singular locus of $X$.

\subsection{Blow-up: turning singularities into smooth points}

Resolution of singularities is a classical central problem in geometry. By resolution we mean that we substitute the original,
possibly singular $X$ with a nonsingular $Y$ with a proper 
birational map $f:Y \to X$ such that $f$ is an isomorphism over some open dense subset of X.

The celebrated Hironaka theorem \cite{hironaka} from 1964 asserts that such resolution exists for all $X$, and it can be constructed as a series of  
elementary operations, called blowing up. Blowing up or blow-up is a type of geometric transformation which replaces a subspace of a given space with all the tangent directions pointing out of that subspace. 

\begin{figure}[ht!]
\centering
\includegraphics[width=70mm]{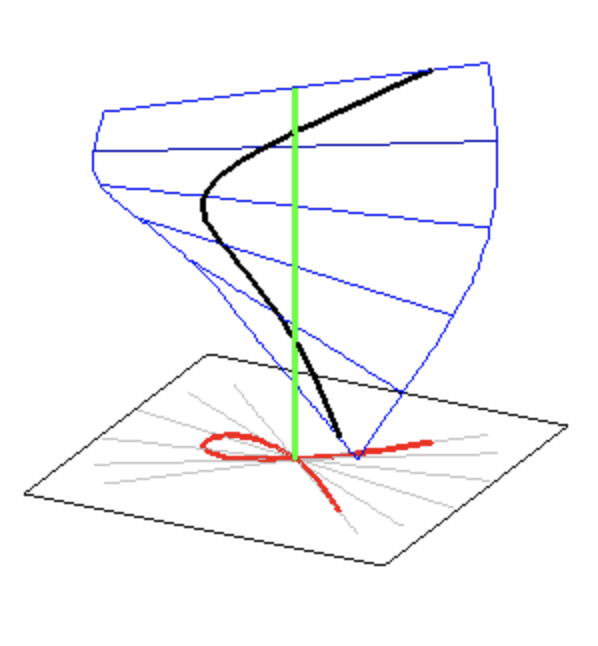}
\caption{Blowing up the nodal red curve $y^2-x^2(x+1)=0$: substitute $y=xt$ to get $x^2(t^2-x-1)=0$, which has the green and black component. The non-singular blown-up curve is the black curve $t^2-x-1=0$}
\label{fig:blowup}
\end{figure}

For example, the blow-up of a point in a plane replaces the point 
with the projectivized tangent space at that point, and this gives a resolution of the nodal curve $y^2-x^2(x+1)=0$. Over the field of real numbers, a picture can be illustrated in Figure \ref{fig:blowup}.

More precisely, Hironaka proved that the resolution of singularities can be achieved by a sequence of blowups 
$$Y=X_n \to X_{n-1} \to \ldots \to X_0=X$$
if the characteristic of the base field is zero.

This beautiful and fundamental work was recognized with a Fields medal in 1970. Villamayor, as well as Bierstone and Milman independently, have described the algorithmic nature of the process of resolving singularities in characteristic zero. Hironaka's theorem has been proven through de Jong's innovative ideas, leading to simple proofs by Abramovich and de Jong, as well as by Bogomolov and Pantvev. The most recent significant development in devising a straightforward resolution algorithm has been made by Abramovich, Tempkin, and Vlodarczyk, as well as by McQuillen, who proposed a simple stacky presentation through the use of weighted blow-ups. 

\section{Solutions to the resolution problem}

In the literature, there appear several proofs for Hironaka's celebrated theorem on the resolution of singularities
of varieties of arbitrary dimension defined over fields of characteristic zero. These proofs associate invariants to 
singularities, and show that certain type of blow-ups improve the invariant. 

We can interpret resolution as a game between two players. Player A attempts to improve the singularities. 
Player B is some malevolent adversary who tries to keep the singularities alive as long as possible. 
The first player chooses the centres of the blowups, the second provides new order functions after each blowup. 

The formulation of the resolution as a game goes back to Hironaka himself. He introduced the polyhedra game where 
Player A has a winning strategy, which provide resolution of hypersurface singularities. He formulated a "hard" 
polyhedra game, where a winning strategy for Player A would imply the resolution theorem in full generality, but such 
winning strategy does not necessarily exist. 
Later Hauser defined a game which provided a new proof of the Hironaka theorem. 

%Certain modified version of the game, due to Bloch and Levine provide solutions to the moving cylcles problem. 
%Finally, recent work of B\'erczi \cite{bercziThom,berczitau} shows that a restricted weighted version of the Hironaka game provides closed integration
%formulas over Hilbert scheme of points on manifolds.  

The most basic version of the game is defined in section \ref{fundamental_hironaka_game}. Here we list some (but not all) known variations with their key features. We introduce the Thom game, which provides formulas for Thom polynomials and integrals over Hilbert scheme of points as explained in section \ref{applications}.

\subsection{Hauser game} 

This version of the Hironaka game was suggested by Hauser \cite{hauser1}. A simple winning strategy was given by Zeillinger \cite{zeillinger}, which gives a resolution process for hypersurfaces singularities. 

\textbf{The rules:}
\begin{itemize}
\item states: A finite set of points $S \subset \mathbb{N}^n$, such that $S$ is the set of vertices of the positive 
convex hull $\Delta=\{S+\mathbb{R}^n_+\}$. 
\item move: The host chooses a subset $I\subset \{1,2,\cdots, n\}$ such that $|I|\geq 2$. The agent chooses a number $i\in I$.
\item state change: Given the pair $(I, i)$ chosen by the host and agent, for $x=(x_1,\cdots,x_n)\in \mathbb Z^n$ we define
$T_{I,i}(x)=(x_1',\ldots, x_n')$ where 
$$x_j' = \begin{cases} x_j, &\qquad\text{if } i\neq j \\ \sum\limits_{k\in I} x_k, &\qquad\text{if }i=j
\end{cases},$$
The new state $S'$ is formed by the vertices of the Newton polyhedron of $\Delta'=\{T_{I,i}(x):x\in S\}$.
\item terminal states: a state $S$ is terminal if it consists of one single point. 
\end{itemize}
In short, the host wants to reduce the size of $S$ as quickly as possible, but the agent wants to keep the size of
$S$ large.

\subsection{Hironaka's polyhedra game} 

This is the original Hironaka game from 1970, see \cite{hironaka}. A winning strategy for the host was given by Mark Spivakovsky in 1980 \cite{spivakovsky2} which proved the resolution theorem for hypersurfaces.  

\textbf{The rules:}
\begin{itemize}
\item states: A finite set of rational points $S \subset \mathbb{Q}^n$, such that $\sum_{i=1}^n x_i>1$ for all 
$(x_1,\ldots, x_n)\in S$, and $S$ is the set of vertices of the positive 
convex hull $\Delta=\{S+\mathbb{R}^n_+\}$. 
\item move: The host chooses a subset $I\subset \{1,2,\cdots, n\}$ such that $|I|\geq 2$ and 
$\sum_{i\in I}x_i\ge 1$ for all $(x_1,\ldots, x_n)\in S$. The agent chooses a number $i\in I$.
\item state change: Given the pair $(I,i)$ chosen by the host and agent, for $x=(x_1,\cdots,x_n)\in \mathbb Z^n$ we define
$T_{I,i}(x)=(x_1',\ldots, x_n')$ where 
$$x_j' = \begin{cases} x_j, &\qquad\text{if } i\neq j \\ \sum\limits_{k\in I} x_k -1, &\qquad\text{if }i=j
\end{cases},$$
The new state $S'$ is formed by the vertices of the Newton polyhedron of $\Delta'=\{T_{I,j}(x):x\in S\}$.
\item terminal states: a state $S$ is terminal if it consists a point $(x_1,\ldots, x_n)$ such that 
$\sum_{i=1}^n x_i \le 1$. 
\end{itemize}

\subsection{Hard polyhedra game} 

The hard polyhedra game was proposed by Hironaka in 1978 \cite{hironaka2}
Hironaka has proved that a winning strategy for the host of this game would imply the local uniformization theorem for an
algebraic variety over an algebraically closed field of any characteristic.
However, a famous result of Mark Spivakovsky \cite{spivakovsky} showed that the host does not always have a winning strategy

\textbf{The rules:}
\begin{itemize}
\item states: A finite set of rational points $S \subset \mathbb{Q}^n$, such that $\sum_{i=1}^n x_i>1$ for all 
$(x_1,\ldots, x_n)\in S$, the denominators are bounded by some fix $N$, and $S$ is the set of vertices of the positive 
convex hull $\Delta=\{S+\mathbb{R}^n_+\}$. 
\item move: The host chooses a subset $I\subset \{1,2,\cdots, n\}$ such that $|I|\geq 2$ and 
$\sum_{i\in I}x_i\ge 1$ for all $(x_1,\ldots, x_n)\in S$. 
The agent chooses some element $i\in S$ and modifies the Newton polygon $\Delta$ to a set $\Delta^*$ by
the following procedure: first, the agent selects a finite number of points $y=(y_1,\ldots, y_n)$, all of whose 
coordinates are rational numbers with denominators bounded by $N$ as above, and for each of which there exists
an $x = (x_1, \ldots, x_n)\in \Delta$ which satisfy some basic relations. $\Delta^*$ is then taken to be the positive 
convex hull of $\Delta \cup \{\text{selected points}\}$.

\item state change: Given the pair $(I,i)$ chosen by the host, for $x=(x_1,\cdots,x_n)\in \mathbb Z^n$ we define
$T_{I,i}(x)=(x_1',\ldots, x_n')$ where 
$$x_j' = \begin{cases}x_j, &\qquad\text{if } i\neq j \\ \sum\limits_{k\in I} x_k -1, &\qquad\text{if }i=j
\end{cases},$$
The new state $S'$ is formed by the vertices of the Newton polyhedron of $\Delta'=\{T_{I,j}(x):x\in S\}$.
\item terminal states: a state $S$ is terminal if it consists a point $(x_1,\ldots, x_n)$ such that 
$\sum_{i=1}^n x_i \le 1$. 
\end{itemize}

\subsection{The Stratify game}

In 2012 Hauser and Schicho \cite{hauser1} introduced a combinatorial game, called Stratify. It exhibits the axiomatic and logical 
structure of the existing proofs for the resolution of singularities of algebraic varieties in characteristic zero. 
The resolution is typically built on a sequence of blowups in smooth centres which are chosen as the smallest stratum 
of a suitable stratification of the variety. The choice of the stratification and the proof of termination of the 
resolution procedure are both established by induction on the ambient dimension. 

\subsection{Thom game}

This is a modified version of the Hironaka game which we developed in the present paper to find optimal solutions for Thom polynomials and integrals over the Hilbert scheme of points as explained in section \ref{applications}.  
The Thom game is a weighted version of the Hironaka game. It has a winning strategy, and
every run of the game provides a blow-up tree, which encodes a formula for Thom polynomials of singularities, answering 
long-standing question in enumerative geometry.

\textbf{The rules:}
\begin{itemize}
\item states: A pair $(S,w)$, where: $S$ is a finite set of points $S \subset \mathbb{N}^n$, such that $S$ is the set of 
vertices of the positive convex hull $\Delta=\{S+\mathbb{R}^n_+\}$; $w=(w_1,\ldots, w_n)\in \mathbb{N}^n$ is a weight 
vector associating a nonnegative integer weight to all coordinates.
\item move: The host chooses a subset $I\subset \{1,2,\cdots, n\}$ such that $|I|\geq 2$ and 
$\sum_{i\in I}x_i\ge 1$ for all $(x_1,\ldots, x_n)\in S$.
The agent chooses an $i\in I$ such that $w_i$ is minimal in $\{w_j: j\in I\}$.
\item state change: Given the pair $(I,i)$ chosen by the host and agent, for $x=(x_1,\cdots,x_n)\in \mathbb Z^n$ we define
$T_{I,i}(x)=(x_1',\ldots, x_n')$ where 
$$x_j' = \begin{cases}x_j, &\qquad\text{if } i\neq j \\ \sum\limits_{k\in I} x_k, &\qquad\text{if } i=j
\end{cases},$$
The new state $S'$ is formed by the vertices of the Newton polyhedron of $\Delta'=\{T_{I,i}(x):x\in S\}$, shifted by
a positive integer multiple of $(-1,\ldots, -1)$ such that $S'$ still sits in the positive quadrant, but any
further shift will move it out. 
The new weight vector is 
$$w'_j=\begin{cases}w_j, &\qquad\text{if } j=i \text{ or } j\notin I \\ w_j-w_i &\qquad\text{if } j \in I\setminus \{i\}
\end{cases},$$
\item terminal states: a state $S$ is terminal if it consists of one single point. 
\end{itemize}

\subsection{The Abramovich-Tempkin-Wlodarczyk game}

In 2020 Abramovich, Tempkin and Wlodarczyk \cite{atv} introduced a new resolution algorithm, based on weighted blow-ups. Quillen \cite{quillen} independently concluded similar results. Their resolution process uses intrinstic invariants of singularities which improves after each blow-up, resulting in a significantly simpler proof of the Hironaka theorem. In their original version, there is no choice in the blowing-up process, and our calculations with the Thom ideals indicate that the ATW resolution can be far from being optimal in terms of the number of leaves of the blowing-up tree. However we are working on transforming the ATW algorithm  into a game with a view towards an ML approach.

\section{A host trained with MCTS}\label{appendix:alphazero}
The host in section \ref{MCTS_planning} is trained using a custom implementation 
%\footnote{see \href{https://github.com/honglu2875/hironaka/tree/main/hironaka/jax}{https://github.com/honglu2875/hironaka/tree/main/hironaka/jax}} 
of AlphaZero, with host and agent being different neural network and trained back-and-forth using MCTS planning taking turn fixing the counter-party. We evaluate the host against the choose-first agent and the choose-last agent (the benchmark agents who always pick the first/last action) and cherry-pick the checkpoint who has the best $\rho$ score against both of them. 

Both host and agent networks use the simplest fully-connected neural network with ReLU as their activation functions. The detailed spec is in Table \ref{model-spec}.

\begin{table}[t]
\caption{model details}
\label{model-spec}
\vskip 0.1in
\begin{center}
\begin{small}
%\begin{sc}
    \begin{tabular}{l|cccc}
    %\backslashbox{role}{agent}
         & number of layers & hidden dimension & batch size & learning rate \\
         \midrule
         host & 8 & 256 & 512 & 0.0001 \\
         agent & 6 & 256 & 512 & 0.0001 \\
         \bottomrule
    \end{tabular}
%\end{sc}
\end{small}
\end{center}
\vskip -0.1in
\end{table}

The rule of the game uses the basic Hironaka game (section \ref{fundamental_hironaka_game}) with one extra transformation at the beginning of each turn: translating all points altogether, so that all coordinates remain non-negative while for each coordinate, at least one point touches the coordinate plane (the coordinate being $0$). In the resolution of hypersurface singularities, this corresponds to removing exceptional divisors and only looking at the strict transforms.

Although our policy does not outperform a guaranteed winning strategy in terms of $\rho$ (see Table \ref{eval}), it is getting close. Note that $\rho^{-1}$ only measures the overall average steps to end a game, which does not reveal other information such as whether it is guaranteed winning or whether the resolution is minimal. 

In the following subsections, we fix our host policy, and demonstrate a few complete state-action trees against all agent choices with initial states corresponding to du Val singularities on surfaces. 

\subsection{Conventions}
Let us first explain some conventions and notation changes.

In this section, we make a minor change of notation: 
\begin{itemize}
    \item the label of coordinates will be $0$-based instead of $1$-based (see the notation of section \ref{fundamental_hironaka_game}). Concretely, host action $I\subset \{0, 1, \cdots, n-1\}$.
\end{itemize}

We would also like to repeat and highlight the additional translation rule during state transition:
\begin{itemize}
    \item after the linear transformation, we post-compose with a translation, so that all coordinates remain non-negative while for each coordinate, there exists at least one point which touches the coordinate plane (the coordinate being $0$)
\end{itemize}

Recall that the during the state transition, the host first chooses a subset $I\subset\{0, \cdots, n-1\}$ and agent chooses a number from $I$. Although the state transition consists of 1) host action, 2) agent action, we note that 
\begin{itemize}
    \item it is only necessary to draw the agent actions as edges, because all possible agent choices will recover the subset $I$.
\end{itemize}
Therefore, in the full trees we will demonstrate, the edges only correspond to agent choice, and they will be labeled by their corresponding coordinate.

\subsection{A2 surface singularity}
We first show the tree of A2 singularity where the host easily found the minimal resolution. We use this simple example to explain in details about how to parse our pictures.
\begin{figure}[ht!]
\centering
\includegraphics[width=80mm]{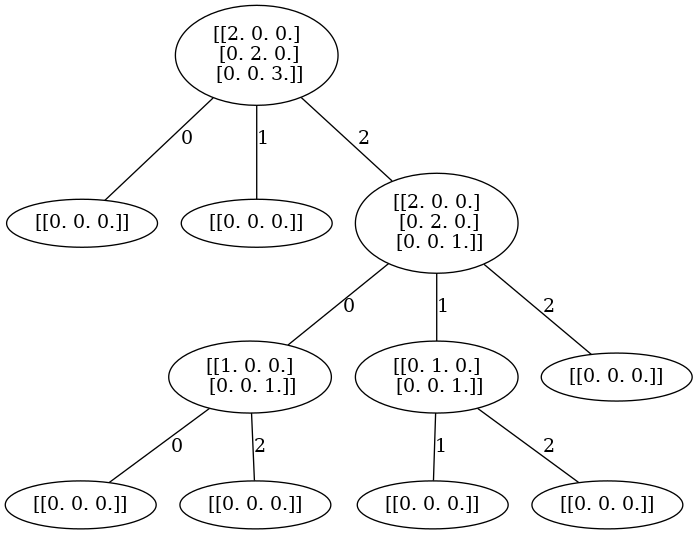}
\caption{The host policy on A2 singularity} 
\end{figure}

We consider affine surfaces defined in $\mathbb C^3$, and name the three coordinates $x,y,z$. The initial state (the root) consists of three points: $(2,0,0), (0,2,0), (0,0,3)$ corresponding to the hypersurface $x^2 + y^2 + z^3 = 0$.

There are $3$ edges coming out of the root and labelled as $0, 1, 2$. This implies that our policy chose the action of $I=\{0,1,2\}$ (according to the notation of \ref{fundamental_hironaka_game}) which corresponds to the blow-up center $x=y=z=0$. The agent now has $3$ charts (actions) to pick from. 

For example, after taking action $0$, according to the rule of the basic Hironaka game plus our additional translation rule, the three points undergo the following operations:
\begin{itemize}
\item becoming $(2,0,0), (2,2,0), (2,0,3)$ after adding the first and the second coordinate to the third coordinate;
\item removing $(2,2,0)$ and $(2,0,3)$ as they are in the interior of the Newton polytope.
\item translating $(2,0,0)$ to $(0,0,0)$ according to our extra rule of translation explained in the beginning of Appendix \ref{appendix:alphazero}.
\end{itemize}
Since we have only one point left, the game terminates at $(0,0,0)$ after the agent took the action $0$.

The only way to continue the game is for the agent to take the action $2$. After this action, the three points first become $(2,0,2), (0,2,2), (0,0,3)$. None of the points are interior points, therefore all of them survive and get translated to $(2,0,0), (0,2,0), (0,0,1)$ along the vector $(0,0,-2)$ so that the first two points have the third coordinate being $0$. 

The next host action is also $I=\{0,1,2\}$. Note that the A2 singularity is already smoothed after this blow-up, while the game can still continue after the agent takes action $0$ or $1$. This is simply because the terminating condition of the basic Hironaka game is merely a sufficient condition of smoothness. For example, the state $(1,0,0), (0,0,1)$ corresponds to the hyperplane $x + z = 0$ which is already smooth. One can easily modify the terminating rule to include this case (e.g. the sum of all coordinates being $1$ for at least one point), but we do not do so for consistency.

As a result, this A2 resolution according to the host policy is the minimal resolution.

For readers not having algebraic geometry background, we encourage them to still continue with the D4 singularity where we prepared a detailed chart-by-chart analysis demonstrating how to parse the blow-up information purely mathematically.

\subsection{D4 surface singularity}
We start with the equation
\[ x^2 + y^2z + z^3 = 0\]
which defines a rational double point of type D4.

\begin{figure}[ht!]
\centering
\includegraphics[width=80mm]{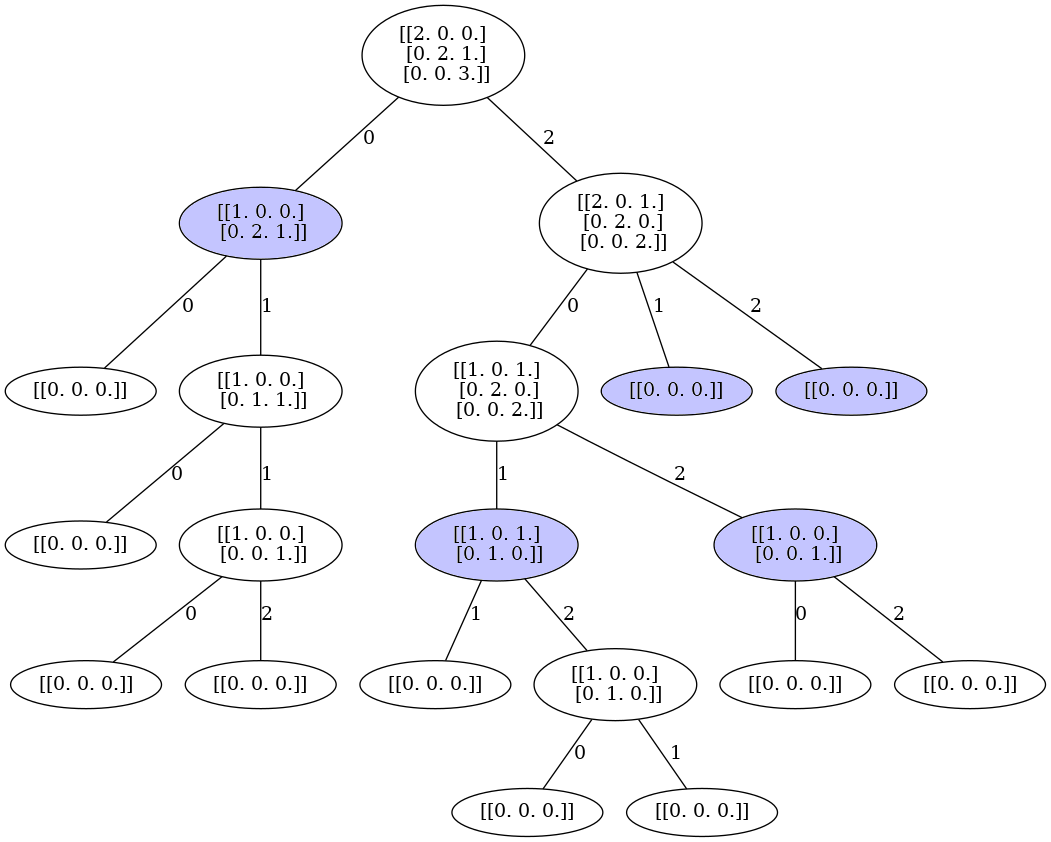}
\caption{The host policy on D4 singularity} \label{d4}
\end{figure}

The host demonstrates a slightly different blow-up path that ended up with the same D4 Dynkin diagram. The mapping between our full policy tree and the RL environment is already explained in the A2 example. So, we use this example to demonstrate a chart-by-chart calculation for readers not coming from algebraic geometry background and still looking to verify the geometry on the mathematical side.

\subsection{A more detailed chart-by-chart analysis}\label{detailed}
Now we are looking at the D4 resolution (Figure \ref{d4}).

Recall that the game may not end even when the chart is smooth. We mark the ealiest smooth states in blue. Throwing away all the sub-trees coming after the blue nodes, we see that it takes two blow-ups to resolve this D4 singularity.

\subsubsection{Step 1, host move}
The host chose the coordinates $I = \{0, 2\}$ as the first step, which corresponds to blowing up the line defined by $x=z=0$. The resulting surface is no longer an affine surface, and to look at the whole picture, one must try to observe through two different charts (think about atlas, charts in manifold theory):
\begin{itemize}
\item $(u,v,w)$ through the change of variables $u=x, v=y, uw=z$.
\item $(u',v',w')$ through $u'w'=x, v'=y, w'=z$.
\end{itemize}
 
Choosing any chart corresponds to an agent action. And the change of variables corresponds to the linear transformations of the game states when looking at the exponents. This step is particularly surprising for me at a first glance, as the usual approach is to blow up $x=y=z=0$ (e.g., the first chart would be $u=x, uv=y, uw=z$, etc). But turns out it doesn't hurt the result.

A smart agent should choose the second chart, as the origin of the first chart is a already smooth point. Let us check this statement by hand: Plugging in $u=x, v=y, uw=z$, we obtain 

$$u^2+uv^2w+u^3w^3=u(u+v^2w+u^2w^3)=0.$$

The equation now defines two surfaces:
\begin{itemize}
    \item $u=0$ which corresponds to the \emph{exceptional divisor} of the blowup of $\mathbb A^3$. It is the "shadow" coming from the modification of the outer space $\mathbb A^3$, and spans outside the surface.
    \item $u+v^2w+u^2w^3=0$ which corresponds to the \emph{strict transform} of the original surface. It is the real modification of the original surface, and it is what we care. One can apply the Jacobian criterion to verify that this is a smooth surface.
\end{itemize}

From the first chart, we can see that the exceptional curves consist of two lines: they are defined by $u+v^2w+u^2w^3=0, u=0$. By plugging $u=0$ in, the system of equation becomes $v^2w=0, u=0$, or equivalently, the line $u=v=0$ unions the line $u=w=0$.

\subsubsection{Step 1, agent move}
The agent chose the second chart. Now by plugging in $u'w'=x, v'=y, w'=z$, we obtain $u'^2w'^2+v'^2w'+w'^3=w'(u'^2w'+v'^2+w'^2)=0$. Again,
\begin{itemize}
    \item $w'=0$ is the exceptional divisor of the blowup on the ambient space $\mathbb A^3$.
    \item $u'^2w'+v'^2+w'^2=0$ is a singular surface. It glues together with $u+v^2w+u^2w^3=0$ and we are just seeing the two parts of the same surface.
\end{itemize}

(As a side note, the two exceptional divisors from different charts $u=0$ and $w'=0$ are not mutually exclusive. They glue together to form one single quasi-projective variety. We are also looking at two parts of the same exceptional divisor. As a result, the exceptional line $u=v=0$ and $w'=v'=0$ are in fact charts of the same $\mathbb P^1$.)

An easy application of Jacobian criterion tells us that the origin $(0,0,0)$ on $u'^2w'+v'^2+w'^2=0$ still needs to be resolved.

\subsubsection{Step 2, host move and agent move}
Now we rinse and repeat from the new equation $u'^2w'+v'^2+w'^2=0$. The host chose all coordinates this time, which corresponds to blowing up at $u'=v'=w'=0$. With the analysis above as well as the help from the blowup tree, we see that only one chart is interesting (agent's choice of coordinate $0$). Altogether, they correspond to the change of variable:

$$u''=u', u''v''=v', u''w''=w'.$$

By plugging in, we see $u''^3w''+u''^2v''^2+u''^2w''^2=u''^2(u''w''+v''^2+w''^2)=0$. Ignoring the exceptional divisor $u''=0$ (for now), we move on to the next singular surface $u''w''+v''^2+w''^2=0$.

The next step can be easily carried out by imitating our previous procedures, and we leave it as an exercise for the interested readers.

Now, if we backtrace all the steps and keep track of the exceptional curves, passing to its dual graph, we will see the famous Dynkin diagram D4.

\subsection{A few more surface singularities}
In addition, we include the full action trees of A3 and D5 as follows. 

\subsubsection{A3}

\begin{figure}[ht!]
\centering
\includegraphics[width=80mm]{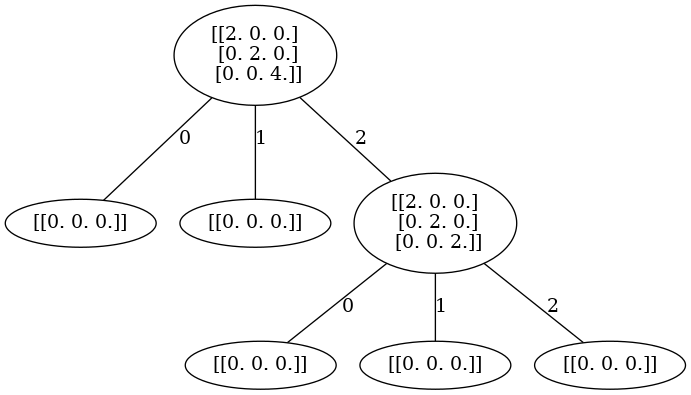}
\caption{The host policy on A3 singularity} \label{a3}
\end{figure}

The host policy tree on A3 is very similar to A2. But the difference is that the first blow-up created two different components in the exceptional locus. From the equation, the agent's choice at coordinate $2$ corresponds to plugging in $x=uz, y=vz$ in $x^2+y^2+z^4=0$. The exceptional locus becomes the intersection of $z=0$ and $u^2+v^2+z^2=0$ and it consists of two complex lines intersecting at the origin.

\subsubsection{D5}
\begin{figure*}[ht!]
\centering
\includegraphics[width=100mm]{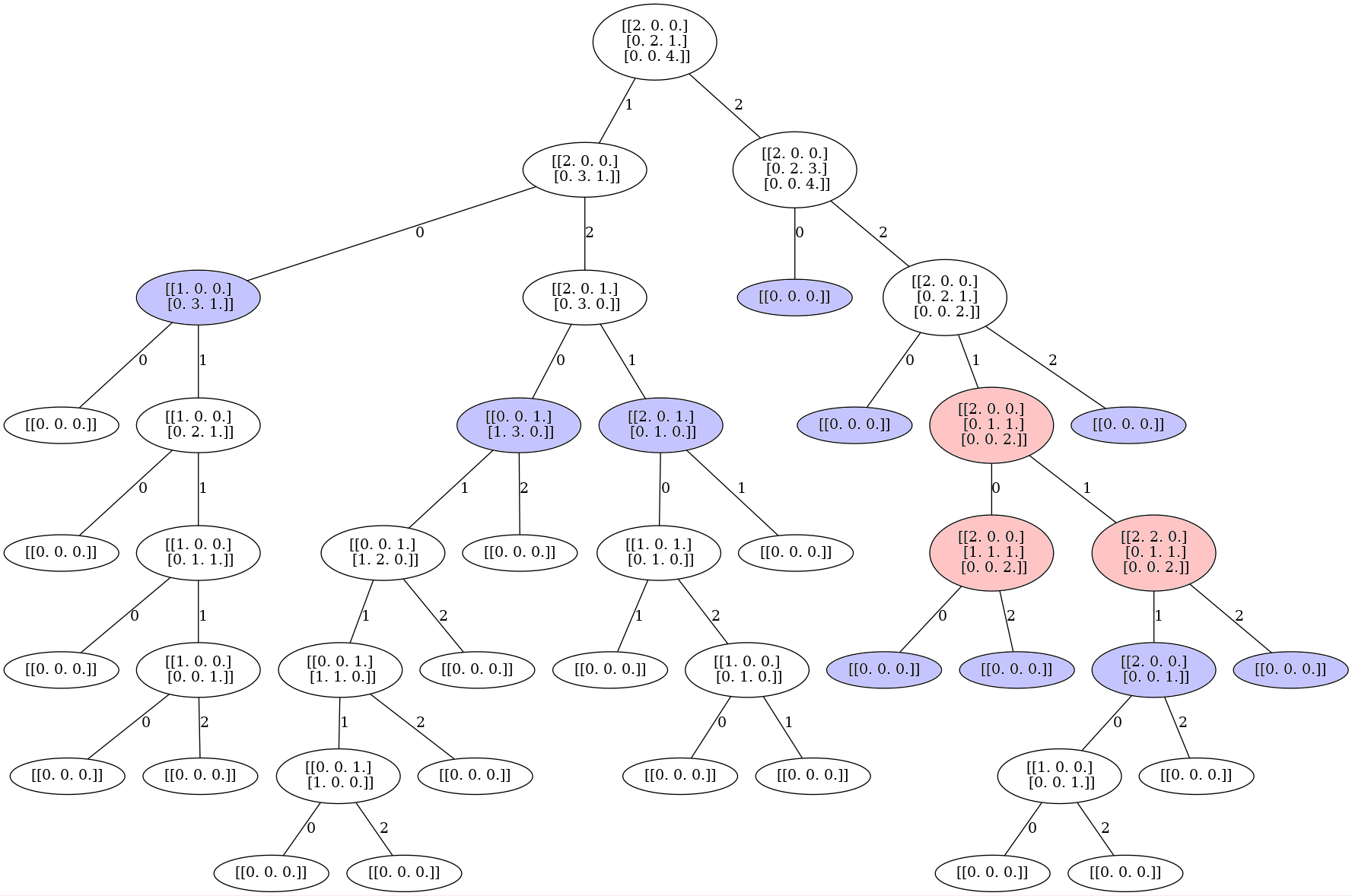}
\caption{The host policy on D5 singularity} \label{d5-host}
\end{figure*}

The full action tree of D5 of our policy is shown in Figure \ref{d5-host}. Note that it is not exactly a minimal resolution, but very close. The host's suboptimal choice is marked in red. Had the host chosen $I=\{0,1,2\}$ for the state transition, it resolves the singularity with $1$ fewer step and recovers the D5 minimal resolution.

\subsubsection{E8}
\begin{figure*}[ht!]
\centering
\includegraphics[width=160mm]{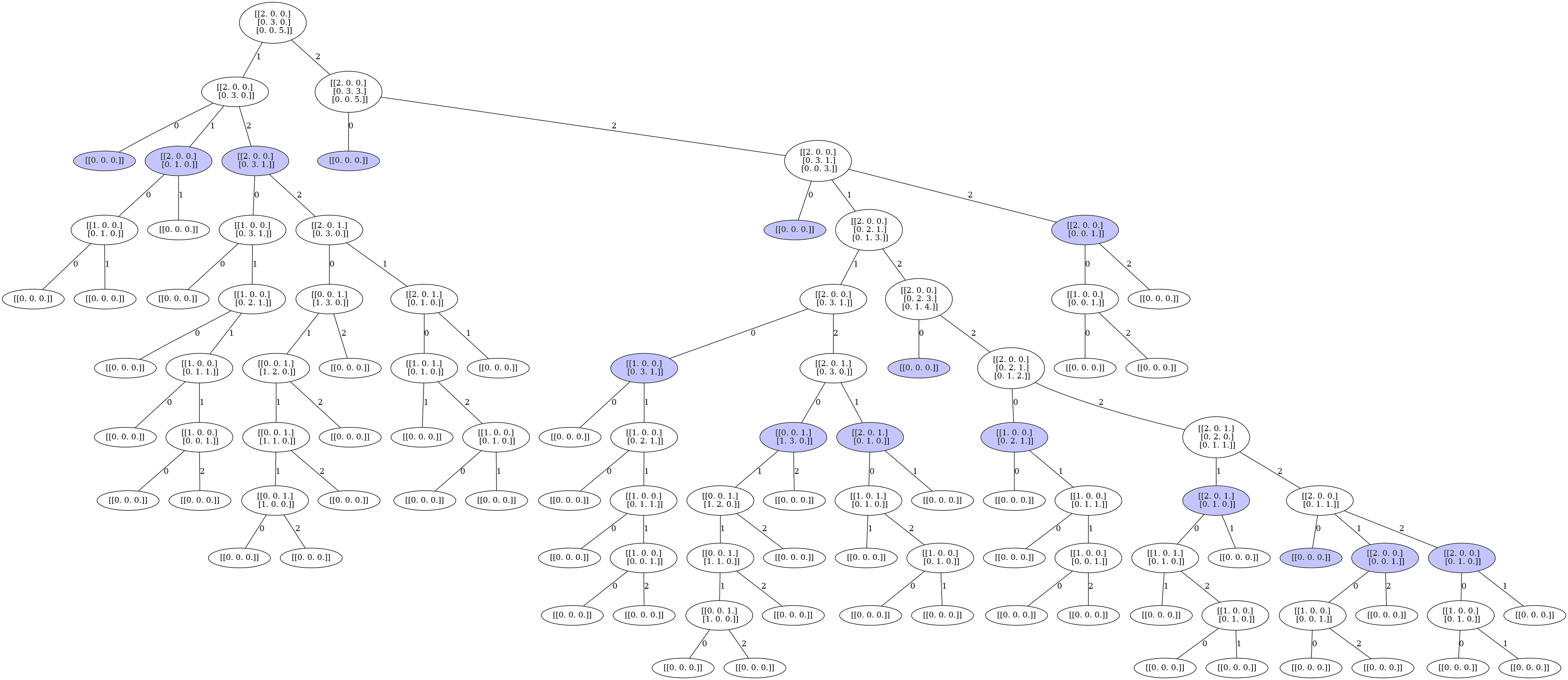}
\caption{The host policy on E8 singularity (blue are the earliest smooth charts where the game do not necessarily terminate)} \label{e8-host}
\end{figure*}

Finally, we include the full action tree of the E8 singularity in Figure \ref{e8-host}. Our host policy was able to terminate the game within $9$ steps against the strongest adversarial agent.

\end{document}